\newif\ifSingleColumn
\SingleColumnfalse
\ifSingleColumn
\documentclass[draftclsnofoot, onecolumn, 12pt]{IEEEtran}
\else
\documentclass[lettersize,journal]{IEEEtran}
\fi
\usepackage{amsmath,amsfonts}
\usepackage{algorithm}
\usepackage{array}
\usepackage{textcomp}
\usepackage{stfloats}
\usepackage{url}
\usepackage{verbatim}
\usepackage{graphicx}
\usepackage{cite}
\usepackage{bm}
\usepackage{amssymb}
\usepackage{algpseudocode} 
\usepackage{algorithmicx}
\usepackage{enumerate}
\usepackage{enumitem}
\usepackage[caption=false,font=normalsize,labelfont=sf,textfont=sf]{subfig}

\newtheorem{proposition}{\bf Proposition}

\newcommand{\rE}{\mathrm{E}}
\newcommand{\rD}{\mathrm{D}}
\newcommand{\rT}{\mathrm{T}}
\newcommand{\rR}{\mathrm{R}}
\newcommand{\btheta}{\bm \theta}
\newcommand{\bphi}{\bm \phi}
\newcommand{\cD}{\mathcal{D}}
\newcommand{\cA}{\mathcal{A}}

\newcommand{\bfun}{\mathbf{f}}
\newcommand{\bgun}{\mathbf{g}}

\newcommand{\rin}{\mathrm{in}}
\newcommand{\rout}{\mathrm{out}}
\newcommand{\bmm}{\bm{m}}
\newcommand{\bmv}{\bm{v}}
\newcommand{\bms}{\bm{s}}
\newcommand{\bmh}{\bm{h}}

\newcommand{\beq}{\begin{equation}}
\newcommand{\eeq}{\end{equation}}

\usepackage{color}

\begin{document}

\ifSingleColumn
\title{\LARGE{Zero-Forget Preservation of Semantic Communication Alignment in Distributed AI Networks}}
\else
\title{\huge{Zero-Forget Preservation of Semantic Communication Alignment in Distributed AI Networks}}
\fi

\author{
\IEEEauthorblockN{
\normalsize{Jingzhi~Hu},~\IEEEmembership{\normalsize Member,~IEEE}
and~\normalsize{Geoffrey Ye Li},~\IEEEmembership{\normalsize Fellow,~IEEE}
}
\thanks{J. Hu and G. Y. Li are with the Department of Electrical and Electronic Engineering, Imperial College London, London SW7 2AZ, UK.}
}

\maketitle

\begin{abstract}
Future communication networks are expected to connect massive distributed artificial intelligence~(AI).
Exploiting \emph{aligned} priori knowledge of AI pairs, it is promising to convert high-dimensional data transmission into highly-compressed semantic communications~(SC).
However, to accommodate the local data distribution and user preferences, AIs generally adapt to different domains, which fundamentally distorts the SC alignment.
In this paper, we propose a zero-forget domain adaptation~(ZFDA) framework to preserve SC alignment. 
To prevent the DA from changing substantial neural parameters of AI, we design sparse additive modifications~(SAM) to the parameters, which can be efficiently stored and switched-off to restore the SC alignment.
To optimize the SAM, we decouple it into tractable continuous variables and a binary mask, and then handle the binary mask by a score-based optimization.
Experimental evaluations on a SC system for image transmissions validate that the proposed framework perfectly preserves the SC alignment with almost no loss of DA performance, even improved in some cases, at a cost of less than 1\% of additional memory.
\end{abstract}


\ifSingleColumn
\newpage
\fi
\section{Introduction}
\label{intro}

From connecting people and things, future communication networks are envisioned to connect massive distributed artificial intelligence~(AI)~\cite{Chen20SUMMIT_Connected} since AIs will become pervasive in society, serving as personal assistants and business enablers~\cite{Tong22WP_6G}.
As the massive number of AIs leads to enormous network traffic, it is imperative for AIs to communicate with optimal efficiency.
Notably, with the powerful feature extraction capabilities enabled by deep learning techniques, AIs can obtain highly-compressed and task-oriented features from raw data.
This makes AI communications inherently suitable for the emerging semantic communications~(SC) paradigm~\cite{Xie21TSP_Deep}.
Following the SC paradigm, the transmitter~(Tx) AI encodes the data into low-dimensional semantics for the receiver~(Rx) AI to decode and complete a target task, so that only the most crucial information is transmitted from Tx to Rx.

The efficacy of SC relies on the prior knowledge of AIs to capture the relationship between data and task.
More importantly, SC requires \emph{alignment} between the knowledge of Tx and Rx AIs~\cite{Luo22WCOM_Semantic}.
As the prior knowledge are embedded in the semantic encoder and decoder, the alignment indicates that the semantic encoder and decoder should work compatibly, reaching a performance level similar to that of being jointly trained.  
To satisfy the requirement of SC alignment, most studies assume the Tx and Rx obtain their knowledge from a shared knowledge base~\cite{Xie21TSP_Deep,Bourtsoulatze19TCCN_Deep,Luo22WCOM_Semantic}.
For AI communications, this assumption seems to be practical since AIs often inherit the same pre-trained neural models to save substantial training costs.
Therefore, when different AIs utilize parts of the same pre-trained neural models as their semantic encoder and decoder, the SC alignment is naturally achieved. 

However, the SC alignment can be easily disrupted when distributed AIs adapt to their local domains.
Here, a \emph{domain} refers to a joint statistic distribution of data and task labels.
Due to the diverse physical locations of devices and preferences of users, distributed AIs encounter various domains and need to adapt to the domains by adjusting their neural parameters.
Although such parameter adjustment in the domain adaptation~(DA) is beneficial for AIs to perform better in their local domains, it may cause serious SC misalignment, incurring significant distortions for the SC between AIs.

To tackle the SC misalignment, existing studies resort to two kinds of approaches, including \emph{tuning-based}~\cite{Zhang23JSAC_Deep,Si24TWC_Post,Xu24COMML_Sync,Choi24TVT_Semantics} and \emph{equalizer-based}~\cite{Sana23GC_Semantic,Fiorellino24Arxiv_Dynamic, Huttebraucker24ISWCS_Soft}.
In~\cite{Zhang23JSAC_Deep}, the authors design an Rx-lead training scheme, where Tx provides the Rx with raw data and task labels, and the Rx tunes its decoder and feedbacks gradients for Tx to tune the encoder.
In~\cite{Si24TWC_Post}, the authors propose to only tune the decoder and design a latent space-based fine-tuning method to circumvent the transmission of raw data. 
In~\cite{Xu24COMML_Sync}, the authors formulate the alignment as a source model estimation problem that can be handled by using semantics on shared raw data.
In~\cite{Choi24TVT_Semantics}, the Tx downloads the Rx's decoder for joint tuning, and reduces the overheads of uploading parameters by only tuning partial decoder.

On the other hand, equalizer-based approaches aim to align the semantic latent spaces of Tx and Rx. 
In~\cite{Sana23GC_Semantic}, the authors partition semantic latent spaces into atom subspaces of different semantic meanings and design a codebook of inter-transformations based on the optimal transport theory.
Later, in~\cite{Huttebraucker24ISWCS_Soft}, the authors improve the equalizer in~\cite{Sana23GC_Semantic} by introducing soft partitioning of the semantic latent space. 
Moreover, in~\cite{Fiorellino24Arxiv_Dynamic}, the authors leverage the similarity between the encoded target data and encoded anchor data to pursue an consistent semantic latent space across varying encoders.

In stark contrast with the existing approaches, which incur large overheads due to the extra joint training or transformations of sematic latent spaces, we address the SC alignment problem at its root.
Inspired by the idea of incremental learning~\cite{Wortsman20NIPS_Super}, we propose a novel framework named zero-forget DA~(ZFDA) to perfectly preserve the SC alignment of AIs when they adapt to different domains.
Our framework enables each AI to achieve DA with sparse additive modifications~(SAM) on its neural model.
For the SAM optimization, we propose an algorithm to represent the SAM as a sparse binary mask multiplied by continuous variables and determine the mask by learning importance scores for neural parameters.

Since the SAM can be stored at a very low cost and easily removed, each AI can swiftly switch between a pre-trained encoder/decoder with perfect SC alignment and a domain-adapted one tailored to its preference.
Therefore, our primary contribution lies in fundamentally eliminating the need for any SC alignment overheads, enabling seamless SC across the network of distributed AIs with diverse preferences.
In the following, we establish a model of SC between AIs in Sec.~\ref{system_model}.
We establish the ZFDA framework in Sec.~\ref{problem} and propose the SAM optimization algorithm in Sec.~\ref{algorithm}.
Experimental results are presented in Sec.~\ref{evaluation}, and a conclusion is drawn in Sec.~\ref{conclusion}.

\section{System Model}\label{system_model}

Without loss of generality, we focus on the SC between a Tx AI and an Rx AI, which are connected by a communication link, either a wireless link in cellar or Wi-Fi networks, or a wired link in backbone networks.
As described in Sec.~\ref{intro}, both AIs inherit the same pre-trained neural model for a certain task from a shared knowledge base.
We assume the neural model has an general encoder-decoder architecture.
By splitting the pre-trained neural model and adopting its parts as the semantic encoder and decoder, the AIs perform highly efficient SC over the link, transmitting semantics rather raw data.

Specifically, the parameters of the pre-trained encoder and decoder are denoted by $\btheta^*\in\mathbb R^{N_{\rE}}$ and $\bphi^* \in\mathbb R^{N_{\rD}}$, respectively, which are trained at the knowledge base by solving
\beq\label{equ_pretrain}
(\btheta^*,\bphi^*) = \arg\min_{\btheta',\bphi'} \sum_{(\bm X,\bm Y)\in \cD}\ell(\bm Y, \bgun_{\bphi'}\circ\bfun_{\btheta'}(\bm X)),
\eeq
where $\circ$ is function composition,
    $\cD$ denotes the pre-training dataset comprising data $\bm X$ and ground-truth task label $\bm Y$, 
    $\ell(\bm Y, \tilde{\bm Y})$ represents the task loss given outcome of neural network $\tilde{\bm Y}$ and truth $\bm Y$,
    and $\bfun_{\btheta}:\bm X\rightarrow \bm S$ and $\bgun_{\bphi}:\bm S\rightarrow \tilde{\bm Y}$ are the encoder and decoder functions parameterized by $\btheta$ and $\bphi$, mapping from $\bm X$ to features, i.e., semantics, $\bm S$ and from $\bm S$ to outcome $\tilde{\bm Y}$, respectively.
As $\btheta^*$ and $\bphi^*$ are obtained by the joint training in~\eqref{equ_pretrain}, they inherently satisfy the SC alignment.

In practice, each distributed AI has a local domain to adapt to, which is dependent on its local data distribution and task outcome preference.
Theoretically, a domain can be modeled as a probability distribution $\Gamma: (\bm X, \bm Y)\rightarrow [0,1]$.
To adapt to their domains, Tx and Rx AIs need to impose changes to $(\btheta^*,\bphi^*)$.
Take Tx AI as an example: it obtains the changes as $(\Delta\btheta_{\rT}^*,\Delta\bphi_{\rT}^*) = \arg\min_{\Delta\btheta, \Delta\bphi} L_{\rT}(\Delta\btheta, \Delta\bphi)$, where
\beq
L_{\rT}(\Delta\btheta, \Delta\bphi)=
\mathbb E_{(\bm X,\bm Y)\sim \Gamma_{\rT}}~\ell_{\rT}(\bm Y, \bgun_{\bphi^*+\Delta\bphi}\circ\bfun_{\btheta^*+\Delta\btheta}(\bm X)). \nonumber
\eeq

Here, $\Gamma_{\rT}$ represents the local domain of Tx AI, $\ell_{\rT}(\cdot)$ denotes the task loss under Tx AI's preference, and $L_{\rT}(\cdot)$ is referred to as the \emph{domain loss}.
Therefore, the \emph{adapted parameters} $(\btheta_{\rT}^*, \bphi_{\rT}^*) \!=\! (\btheta^*,\bphi^*) \!+\! (\Delta\btheta_{\rT}^*,\Delta\bphi_{\rT}^*)$.
Similarly, $\Gamma_{\rR}$, $L_{\rR}(\cdot)$, and $(\btheta_{\rR}^*, \bphi_{\rR}^*)$ are defined for the Rx AI.

\begin{figure}[t] 
    \centering
    \ifSingleColumn
    \includegraphics[width=.6\linewidth]{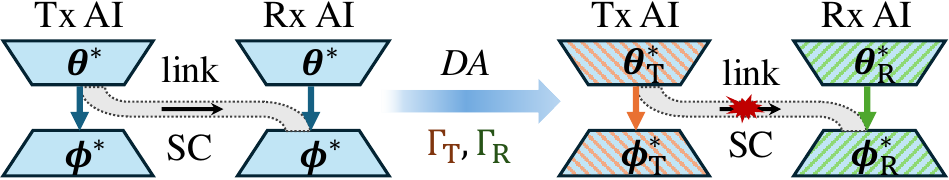}
    \else
    \includegraphics[width=.87\linewidth]{figures/fig_sysmod_clip.pdf}
    \fi
    \caption{SC alignment between Tx and Rx AIs using pre-trained parameters (\textbf{left}); Distorted SC alignment due to respective DA of AIs (\textbf{right}).}
    \label{fig_sys_mod}
\vspace{-.5em}
\end{figure}

Due to the large size of the neural model and the limited local memory, the AIs cannot store a copy of the pre-trained parameters along with the adapted ones. 
As shown in Fig.~\ref{fig_sys_mod}, the AIs need to use adapted semantic encoder and decoder for the SC, which suffers from misalignment causing increase of task loss.
Assume that the SC between the Tx and Rx AIs aims to transmit data in the same distribution of pre-training, then the \emph{misalignment loss} can be modeled as
\beq\label{equ_increase_loss}
J(\btheta_T^*,\bphi_R^*) = L(\btheta_T^*,\bphi_R^*) - L(\btheta^*,\bphi^*),
\eeq
where $L(\btheta,\bphi) = \frac{1}{|\cD|}\sum_{(\bm X,\bm Y)\in \cD} \ell(\bm Y, \bgun_{\bphi}\circ\bfun_{\btheta}(\bm X))$.
To re-align their semantic encoder and decoder and reduce the misalignment loss, the Tx and Rx AI perform an \emph{alignment process}, which can be denoted by function $\cA:(\btheta_{\rT}^*,\bphi_{\rR}^*)\rightarrow(\btheta'_{\rT},\bphi'_{\rR})$, so that $J(\cA(\btheta_{\rT}^*,\bphi_{\rR}^*))<J(\btheta_T^*,\bphi_R^*)$.

In \eqref{equ_increase_loss}, we implicitly model the communication link as an identify function for the simplicity of presentation.
This is based on that the semantics can be reliably transmitted as bits by the modern mobile networks, which can provide above $99.9\%$ reliability according to the 3GPP standard~\cite{ETSI_TS_122_261_2019}.
Although this may lose a little efficiency compared to the joint semantic and channel coding~(JSCC) scheme~\cite{Bourtsoulatze19TCCN_Deep}, where raw data are directly mapped to symbols on lossy wireless channels, it is more compatible to general infrastructure.

\section{ZFDA Framework to Preserve SC Alignment}\label{problem}

Our goal of achieving the SC alignment for the Tx and Rx AIs, with their respective parameters been adapted to different domains, can be formulated as an optimization problem of $\cA$:
\begin{align}
    \text{(P0)}\quad \min_{\cA}~&J(\cA(\btheta_{\rT}^*+\Delta\btheta_{\rT}^*,\bphi_{\rR}^*+\Delta\bphi_{\rR}^*)). \nonumber 
\end{align}

However, it is a major challenge to solve (P0) efficiently.
Firstly, since $\cA$ is a functional variable, it can take arbitrary form.
This vast degree of optimization freedom makes it intractable to solve by conventional methods.
Secondly, for $\cA$ to be a practical solution, it should introduce as little overhead as possible.
Without consideration of efficiency, several trivial solutions of (P0) can be readily identified, including jointly re-training the encoder and decoder, or naively memorizing and restoring the original $(\btheta^*,\bphi^*)$.
Unfortunately, they either incur large overhead in computation and communications, or impose a heavy burden on memory.
Although equalizer-based approaches~\cite{Sana23GC_Semantic,Fiorellino24Arxiv_Dynamic, Huttebraucker24ISWCS_Soft} seems to reduce such burden, it remains a challenging problem to find the optimal transformations between semantic latent spaces of the Tx and Rx AIs.

To address (P0) efficiently, we propose a novel SC alignment framework named ZFDA, approaching the problem from a new perspective. 
Instead of relying on post-hoc remedies, ZFDA leads to adapted parameters that are easy to restore the original alignment without any forgetting.
More specifically, it uses sparse additive modifications~(SAM) to achieve the DA, which can be stored at a low cost and switched-off efficiently.
As the neural parameters can be restored to its pre-trained state, the misalignment loss is reduced to zero.
In this regard, we convert (P0) to finding the SAM for DA.
Again, take the Tx AI as an example: solving the SAM can be formulated as\footnote{
    In (P1)$_{\rT}$, we can relax constraint~\eqref{equ_sparse_constraint} by only requiring the modifications to the encoder to be sparse. Although this benefits the domain loss minimization, it essentially assumes that the Tx AI will not be at the Rx end in the future.
}
\begin{align}
\text{(P1)$_{\rT}$}~\min_{\Delta{\btheta}, \Delta{\bphi}}~&L_{\rT}(\Delta{\btheta}, \Delta{\bphi}) \nonumber\\
\text{s.t.}~& \|\Delta\btheta\|_0 + \|\Delta\bphi\|_0 \leq \gamma\cdot(N_{\rE}+N_{\rD}), \label{equ_sparse_constraint}
\end{align}
where $0 \!<\! \gamma\!\ll\! 1$ represents the \emph{sparsity ratio}, and $\|\cdot\|_0$ denotes the $\ell_0$-norm counting non-zero elements.
Similarly, problem (P1)$_{\rR}$ can be formulated for the Rx AI.
By solving (P1)$_{\rT}$ and (P1)$_{\rR}$, optimized SAMs for the encoder and decoder can be obtained, denoted by $\Delta\hat{\btheta}^*_{\rT}$ and $\Delta\hat{\bphi}^*_{\rR}$, respectively.
Therefore, denote the adapted parameters in the ZFDA framework by $\hat{\btheta}^*_{\rT}$ and $\hat{\bphi}_{\rR}^*$, and the optimal alignment process:
\beq
\cA^*(\hat{\btheta}^*_{\rT}, \hat{\bphi}_{\rR}^*) = (\hat{\btheta}^*_{\rT} - \Delta\hat{\btheta}_{\rT}^*, \hat{\bphi}_{\rR}^* - \Delta\hat{\bphi}_{\rR}^*)=(\btheta^*, \bphi^*).
\eeq

Nevertheless, (P1)${\rT}$ and (P1)${\rR}$ are a challenging variant of combinatorial optimization, aimed at selecting an optimal subset of parameters to modify while simultaneously optimizing the specific adjustments to these parameters. 
Given the vast number of potential subsets, such combinatorial optimization problems are notoriously NP-hard, presenting substantial challenges for efficient resolution.
 
\section{SAM Optimization Algorithm}\label{algorithm}

To efficiently address the problems (P1)$_{\rT}$ and (P1)$_{\rR}$, we propose an SAM optimization algorithm. This algorithm consists of three main components, which are detailed below and summarized in Algorithm~\ref{alg_overall}.
We note that given the strong symmetry between (P1)$_{\rT}$ and (P1)$_{\rR}$, we focus on handling (P1)$_{\rT}$, with the same procedures applicable to (P1)$_{\rR}$.

\subsection{Sparsity-Aware SAM Decomposition}
The first difficulty in solving (P1)$_{\rT}$ is to ensure the sparsity ratio of $\Delta \btheta$ and $\Delta \bphi$ in~\eqref{equ_sparse_constraint}.
To handle this difficulty, we decompose each of them into two components: a binary mask and a continuous vector, i.e., 
\begin{equation}
\label{equ_sam_decompose}
\Delta\btheta = \bmm_{\btheta}\odot \bmv_{\btheta},~
\Delta\bphi = \bmm_{\bphi} \odot \bmv_{\bphi},
\end{equation}
where $\odot$ denotes element-wise product,
$\bmm_{\btheta}\in \mathbb B^{N_{\rE}}$ and $\bmm_{\bphi}\in \mathbb B^{N_{\rD}}$ are binary mask for $\btheta$ and $\bphi$ respectively, 
and $\bmv_{\btheta}\in\mathbb R^{N_{\rE}}$ and $\bmv_{\bphi}\in\mathbb R^{N_{\rD}}$ represent additive modifications to $\btheta$ and $\bphi$ before applying the masks.

Furthermore, to avoid directly restricting the $\ell_0$-norm of the masks, which is difficult to enforce, we introduce an auxiliary variable named importance score, or \emph{score} in short, for each parameter element as in~\cite{Ramanujan20CVPR_What}.
Given sparsity ratio $\gamma$, the score vectors, i.e., $\bms_{\btheta}\in\mathbb R^{N_{\rE}}$ and $\bms_{\bphi}\in\mathbb R^{N_{\rD}}$, determine their respective masks by setting the mask elements with the top-$\gamma$ scores to be one and others to be zero.
For example, consider a score vector $\bm s\in\mathbb R^{N'}$ and its corresponding mask $\bmm\in\mathbb B^{N'}$. 
The relationship between $\bms$ and the $i$-th element of $\bmm$ can be expressed by an indicator function below:
\beq\label{equ_mask}
[\bmm]_i = [\bm h(\bms;\gamma)]_i = \begin{cases}
1,~\text{if $[\bms]_i$ among top-$\gamma$ in $\bms$,}\\
0,~\text{otherwise.}
\end{cases}
\eeq
Therefore, with an awareness of sparsity constraint~\eqref{equ_sparse_constraint}, $\Delta\btheta$ and $\Delta\bphi$ can be expressed as 
\beq
\label{equ_score_decompose}
\Delta\btheta = \bm h(\bms_{\btheta};\gamma_{\rE})\odot \bmv_{\btheta},~
\Delta\bphi = \bm h(\bms_{\bphi};\gamma_{\rD})\odot \bmv_{\bphi},
\eeq
where $\gamma_{\rE},\gamma_{\rD}\in[0,\gamma]$ denote the sparsity ratios for the encoder and decoder, respectively, with $\gamma_{\rE}N_{\rE}+\gamma_{\rD}N_{\rD} = \gamma\cdot(N_{\rE}+N_{\rD})$.
By the decomposition in~\eqref{equ_score_decompose}, we encapsulate the discrete, combinatorial nature of SAM into indication functions $\bm h(\cdot;\gamma_{\rE})$ and $\bm h(\cdot;\gamma_{\rD})$ while guaranteeing the sparsity ratio, facilitating efficient optimization of the SAM.

\begin{figure}[!b]
\vspace{-0.7em}
\begin{algorithm}[H]
\ifSingleColumn
\else
\small
\fi
\caption{SAM Optimization Algorithm}
\label{alg_overall}
\begin{algorithmic} [1]
    \State Calculate the sparsity ratios for the $K$ layers of the neural model as $\gamma_{1},\dots,\gamma_{K}$ based on~\eqref{equ_spare_distribute}.
    \State For each layer $k$, initialize decomposed SAM by $\bms_k$ following the normal distribution, $\bmv_k=\bm 0$, and $\bmm_{k}=\bmh(\bms_k;\gamma_k)$.
    \State Repeat sampling a batch of data-label pairs from domain $\Gamma$ and updating $\bms_k,\bmv_k, \forall k$ by~\eqref{equ_update_v} and~\eqref{equ_s_update} until convergence.
    \State Obtain optimized SAM $(\Delta \btheta^*,\Delta \bphi^*)$ based on~\eqref{equ_sam_decompose}.
\end{algorithmic}
\end{algorithm}
\vspace{-1.em}
\end{figure}

\subsection{Score-Aided SAM Optimization}
Based on~\eqref{equ_score_decompose}, given $\gamma_{\rE}$ and $\gamma_{\rD}$, optimizing the SAM is equivalent to optimizing $\bms_{\btheta}$, $\bms_{\bphi}$, $\bmv_{\btheta}$, and $\bmv_{\bphi}$.
Intuitively, the most prevalent and effective approaches to optimize variables in neural models are the gradient descent based optimizers, such as the widely recognized stochastic gradient descent~(SGD) and Adam.
In such algorithms, variables are updated iteratively.
In particular, iterative update of $\bmv_{\btheta}$ and $\bmv_{\bphi}$ can be expressed as
\beq
\label{equ_update_v}
\bmv_{\btheta}\leftarrow \bmv_{\btheta} - \alpha_{\bm v}\cdot \frac{\partial L_{\rT}}{\partial \bmv_{\btheta}},~\bmv_{\bphi}\leftarrow \bmv_{\bphi} - \alpha_{\bm v}\cdot \frac{\partial L_{\rT}}{\partial \bmv_{\bphi}},
\eeq
where $\alpha_{\bmv}$ represents the learning rate for $\bmv_{\btheta}$ and $\bmv_{\bphi}$.

However, it is intractable to evaluate the gradients of $L_{\rT}$ with respect to~(w.r.t.) $\bms_{\btheta}$ and $\bms_{\bphi}$, as the gradient of $\bm h(\cdot;\gamma)$ is zero almost everywhere.
To tackle this difficulty, we leverage the straight-through estimation method~\cite{Ramanujan20CVPR_What}, which is simple to implement and more effective than sophisticated methods as demonstrated in~\cite{Bengio13NIPS_Estimating}.
In this method, the indicator function $\bm h(\cdot;\gamma)$ is treated as an identity function, allowing gradient calculation to directly pass through it.
For example, for the $i$-th element of $\bm s_{\btheta}$, its gradient is approximated by
\beq\label{equ_st_est}
\frac{\partial L_{\rT}}{\partial [\bms_{\btheta}]_i} \approx \frac{\partial L_{\rT}}{\partial [\bmh(\bms_{\btheta};\gamma_{\rE})]_i} = \frac{\partial L_{\rT}}{\partial [\bmm_{\btheta}]_i}.
\eeq
Therefore, the update for $\bm s_{\btheta}$ and $\bm s_{\bphi}$ can be expressed as 
\beq\label{equ_s_update}
\bms_{\btheta} \leftarrow \bms_{\btheta} - \alpha_{\bm s}\frac{\partial L_{\rT}}{\partial \bmm_{\btheta}},~
\bms_{\bphi} \leftarrow \bms_{\bphi} - \alpha_{\bm s}\frac{\partial L_{\rT}}{\partial \bmm_{\bphi}}.
\eeq

The principle behind~\eqref{equ_st_est} can be explained as follows: If the loss gradient w.r.t. a mask element with a value of one is positive, its score should be updated in the negative direction to decrease the mask element to zero, and vice versa.
As scores change, the mask changes, and the subset of selected parameters changes accordingly.
In~\cite[Thm.~1]{Ramanujan20CVPR_What}, it is proven that for a multi-layer perceptron, score update by~\eqref{equ_s_update} will reduce loss.
We extend that theorem to general neural models with an arbitrary architecture in the following Proposition~\ref{thm_loss_decrease}.

\begin{proposition}
\label{thm_loss_decrease}
\textit{For a neural model with an arbitrary architecture, decompose its SAM into a mask, scores, and additive modification as in~\eqref{equ_mask} and~\eqref{equ_score_decompose}. 
Given fixed modifications, when scores are updated based on~\eqref{equ_s_update}, the update of the mask by~\eqref{equ_mask} monotonically reduces the objective loss.}
\end{proposition}
\begin{IEEEproof}
Please refer to Appendix~\ref{proof_1}.
\end{IEEEproof}

\subsection{Sparsity Distribution Regularization}

The designed approach in~\eqref{equ_mask} and~\eqref{equ_s_update} narrows down the search space of the masks from the complete set of $2^{N}$ ($N=N_{\rE}+N_{\rD})$ possible subsets to the subsets of a fixed size $\gamma N$.
Nevertheless, directly optimizing the masks for all the $N$ parameters as a whole still requires searching over a binary space of size $C_{N}^{\gamma N}$, which is prohibitively large and thus severely hinders effective optimization.
To handle this difficulty, we design a regularization scheme to control the distribution of the sparsity throughout the neural model.

Without loss of generality, assume the neural network has a layered architecture.
We aim to regularize layer-wise sparsity ratios, which is equivalent to allocating a parameter budget to each layer. 
A straightforward scheme is to set uniform sparsity ratios for all the layers, resulting in each layer's parameter budget proportional to its number of parameters.
However, this allocation scheme tends to be heavily biased.
In particular, for the $k$-th layer ($k=1,\dots,K$) with input and output dimensions $d_{\rin,k}$ and $d_{\rout,k}$, respectively, its number of parameters $p_k$ is proportional to $d_{\rin,k}\cdot d_{\rout,k}$ due to the number of connection weights.
Thus, $p_k$ is quadratic in $d_{\rin,k}$ and $d_{\rout,k}$, resulting in large proportion of parameter budget allocated to layers with high dimensional input and output.

To alleviate this bias,  we allocate the parameter budget linearly proportional to the sum of $d_{\rin,k}$ and $d_{\rout,k}$, which is proven to benefit sparse neural models~\cite{Evci20ICML_Rigging}.
Consequently, the regularized sparsity ratio for layer $k$ can be derived as
\beq
\label{equ_spare_distribute}
\gamma_k = \gamma\cdot\frac{d_{\rin,k}+d_{\rout,k}}{\sum_{k'}^K (d_{\rin,k'}+d_{\rout,k'})}\cdot \frac{N}{p_k}.
\eeq

\section{Experimental Evaluations}\label{evaluation}

To validate the ZFDA framework, we implement a SC system for image transmission between AIs.
The neural model equipped by the AIs is a typical autoencoder for image reconstruction/denoising tasks, comprising an encoder with 6 convolutional layers and a decoder with 6 deconvolutional layers, as well as other auxiliary layers for activation and normalization.
The encoder is capable of compressing a $32\!\times\!32\!\times\! 3$ image data to $512$-dim semantics, achieving a bandwidth ratio of $0.167$, and the decoder reconstructs the original image using the semantics.
The pre-training of the autoencoder minimizes the mean squared errors~(MSE) between input and output images of 80 classes of images in the CIFAR100 dataset~\cite{CIFAR100}, which takes 40 epochs with a learning rate of $0.01$ to achieve a peak-SNR\footnote{PSNR measures the ratio between the squared maximum pixel value and the MSE. It is commonly adopted to indicate image reconstruction quality.} (PSNR) of $30.51$~\!dB.
Consequently, when the Tx and Rx AIs respectively use the pre-trained encoder and decoder for the SC, the PSNR of the received image is around $30.51$~\!dB, indicating the perfect SC alignment.
We note that the results in the evaluations are obtained on the test datasets of CIFAR-100, which are not used during training.

\begin{figure}[t] 
    \centering
    \ifSingleColumn
    \includegraphics[width=.64\linewidth]{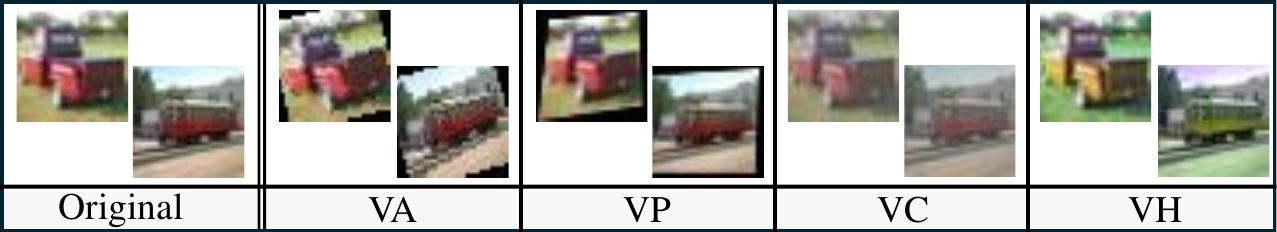}
    \else
    \includegraphics[width=.96\linewidth]{figures/fig_domain_illustration.pdf}
    \fi
    \caption{Samples from the original domain and the other varied local domains.}
    \label{fig_illu_domain}
\vspace{-.5em}
\end{figure}

For each AI, its local domain is represented by a domain dataset comprising of 10 classes of images unused in the pre-training.
Besides, to emulate diverse image distributions in practice, a domain dataset undergoes one of the four variations: varied angle~(VA), varied perspective~(VP), varied contrast~(VC), and varied hue~(VH).
Samples of the original and the varied domains are shown in Fig.~\ref{fig_illu_domain}.
During the DA, an AI starts from the pre-trained parameters and trains them on its domain dataset for 10 epochs with a low learning rate of $10^{-4}$ to avoid over-fitting.

\begin{figure}[b]
    \vspace{-.5em}
    \centering 
    \subfloat{%
    \ifSingleColumn
    \includegraphics[width=0.3\linewidth]{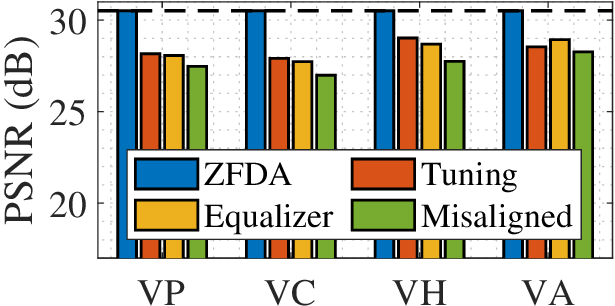}  
    \else
    \includegraphics[width=0.48\linewidth]{figures/exp_1_align_compare_Tx.eps}  
    \fi
    \label{fig_exp1_a}
    }%
    \subfloat{%
    \ifSingleColumn
    \includegraphics[width=0.3\linewidth]{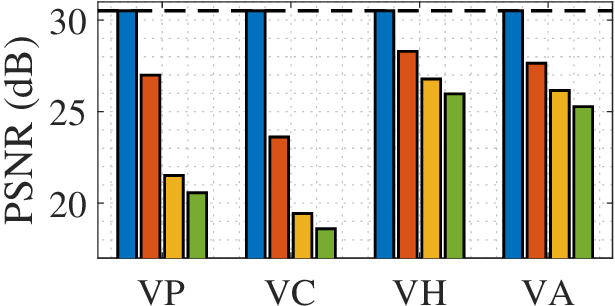}  
    \else
    \includegraphics[width=0.48\linewidth]{figures/exp_1_align_compare_Rx.eps}  
    \fi
    \label{fig_exp1_b}
    }
    \vspace{-.3ex}
    \caption{PSNR for the SC given Tx AI (\textbf{left}) or Rx AI (\textbf{right}) adapted to each varied domain with different alignment methods being employed.}
    \label{fig_exp1}
\end{figure} 

In the left and right parts of Fig.~\ref{fig_exp1}, the misalignment caused by the DA is evaluated when either the Tx or Rx AI adapts to a varied domain. 
In the \emph{misaligned} case, the average PSNR of received images in the original domain significantly drops by 5.34~\!dB in average.
Notably, the DA of Rx decoder results in a larger PSNR loss, 7.91~\!dB in average, which is 5.02~\!dB more than that for Tx encoder.
In re-aligning the SC, we compare the ZFDA with the tuning and equalizer methods.
For the tuning method, we perform the Rx-lead joint training in~\cite{Zhang23JSAC_Deep} for 8 iterations, using 1024 data in total.
For the equalizer method, instead of solving a costly Bayes optimization in~\cite{Sana23GC_Semantic}, we approximate the equalization by using a two-layer dense neural model trained by 1024 data for 30 epochs.
As shown in Fig.~\ref{fig_exp1}, although lots of resources are devoted to SC alignment, the achievable performance is still unsatisfactory: 2.41~\!dB and 0.80~\!dB average increase in PSNR contributed by tuning-based and equalizer-based methods, respectively.
In stark contrast, our ZFDA framework suffers from zero loss with no overheads for gaining the SC alignment as SAMs can be switched-off.

\begin{figure}[t]
    \vspace{-1em}
    \centering 
    \subfloat{%
    \ifSingleColumn
    \includegraphics[width=0.31\linewidth]{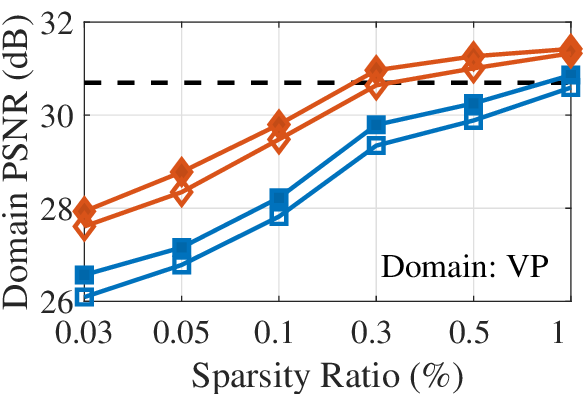}  
    \else
    \includegraphics[width=0.44\linewidth]{figures/fig_Perspect_zfda_effect.eps}  
    \fi
    \label{fig_exp2_a}
    }%
    \subfloat{%
    \ifSingleColumn
    \includegraphics[width=0.32\linewidth]{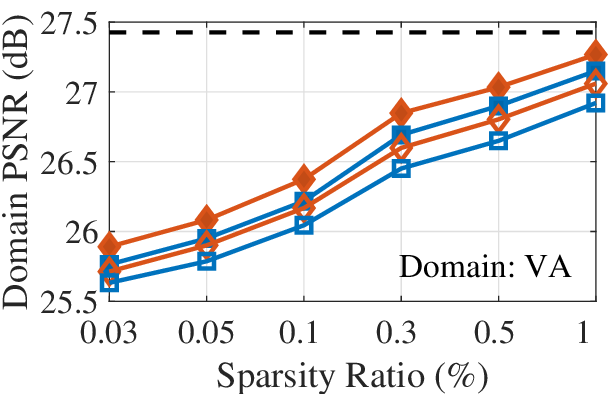}  
    \else
    \includegraphics[width=0.45\linewidth]{figures/fig_Angle_zfda_effect.eps}  
    \fi
    \label{fig_exp2_b}
    }\\
    \subfloat{%
    \ifSingleColumn
    \includegraphics[width=0.31\linewidth]{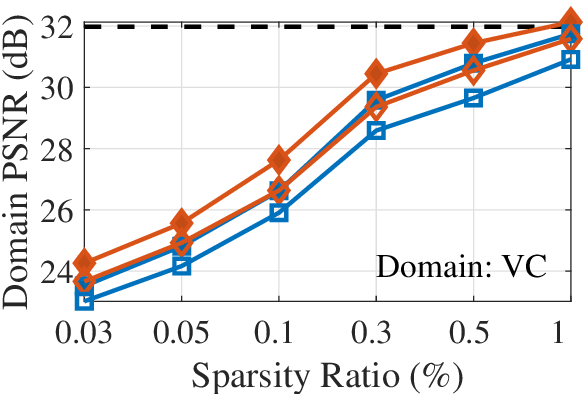}  
    \else
    \includegraphics[width=0.44\linewidth]{figures/fig_Contrast_zfda_effect.eps}  
    \fi
    \label{fig_exp2_c}
    }
    \subfloat{%
    \ifSingleColumn
    \includegraphics[width=0.32\linewidth]{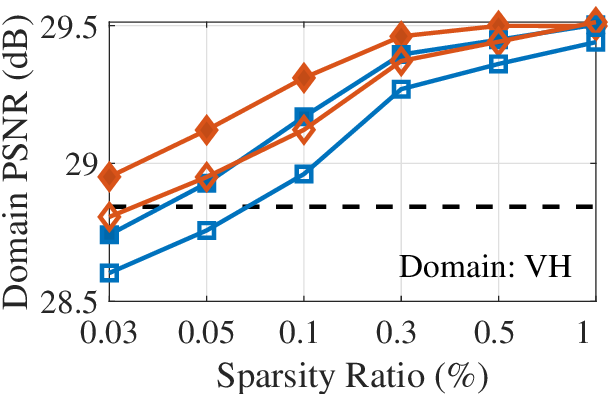}  
    \else
    \includegraphics[width=0.45\linewidth]{figures/fig_Hue_zfda_effect.eps}  
    \fi
    \label{fig_exp2_d}
    }\\
    \subfloat{%
    \hspace{.8em}
    \ifSingleColumn
    \includegraphics[width=0.52\linewidth]{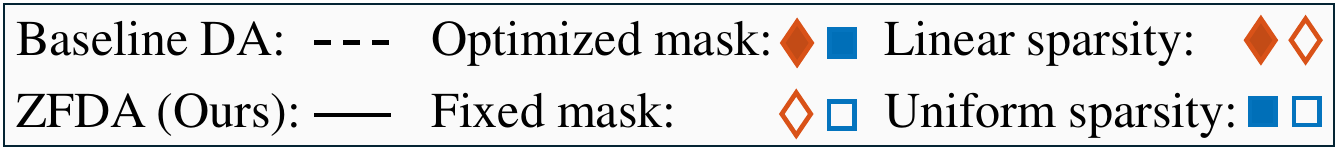}  
    \else
    \includegraphics[width=0.82\linewidth]{figures/fig_legend.pdf}  
    \fi
    \label{fig_exp2_e}
    }
    \vspace{-.7ex}
    \caption{PSNR for the four local domains versus the sparsity rate of the SAM, given different cases of sparsity distribution and mask optimization.}
    \vspace{-.5em}
    \label{fig_exp2}
\end{figure}

A natural question to follow is how ZFDA performs in adapting AIs to local domains.
To address this, we compare the ZFDA with baseline DA in Fig.~\ref{fig_exp2}, which shows the resulting PSNR in the VP, VA, VC, and VH domains given different sparsity ratios $\gamma\in[0.03\%,1\%]$ of ZFDA.
As ZFDA inherently prevents over-fitting of the pre-trained models, we train the SAM on the domain dataset for 30 epochs with $\alpha_{\bms}=1$ and $\alpha_{\bmv}=0.0001$. 
With a sparsity rate of no more than $1\%$, the optimized SAM achieves performance comparable to, or even surpassing, that of baseline DA.
The reason that the ZFDA can outperform the baseline DA is because the SAM limits the number of parameters to change, which helps mitigate over-fitting.
To further validate the SAM optimization, we evaluate four cases, conditioning on whether the mask is optimized by~\eqref{equ_s_update} or fixed after being sampled, and whether the sparsity distribution is a linear one in~\eqref{equ_spare_distribute} or is a uniform one $\gamma_k \!=\!\gamma$.
Fig.~\ref{fig_exp2} shows that the optimized mask and the linear sparsity distribution leads to 0.73~\!dB (18.3\%) and 1.13~\!dB (29.7\%) increase in domain PSNR, respectively.

\section{Conclusion}\label{conclusion}
This paper has proposed a ZFDA framework to preserve the SC alignment for distributed AI networks.
Considering that AIs inherit the same pre-trained neural models, we have analyzed the disruption of their SC alignment due to local DA.
To avoid this disruption, ZFDA achieves the DA by an optimized SAM, which can be efficiently stored and switched off to restore the SC alignment.
Experiment results on practical image transmissions have demonstrated that ZFDA achieves comparable DA performance while allowing zero-forget preservation of the SC alignment, with a storage cost for the SAM of less than 1\% of the neural model's size.

\begin{appendices}
\section{Proof of Proposition~\ref{thm_loss_decrease}}
\label{proof_1}
Without loss of generality, denote the mask and the score vector before score update by $\bmm$ and $\bms$, respectively, and those after the score update by $\tilde{\bmm}$ and $\tilde{\bms}$.
From $\bmm$ to $\tilde{\bmm}$, suppose there are $M$ mask element change from zero to one, with indices $i_{1},\dots,i_M$.
Based on~\eqref{equ_mask}, there should be $M$ mask elements change from one to zero, indexed by $j_{1},\dots,j_M$.
Using the first-order Taylor expansion, the difference between the loss after and before the score update can be calculated by
\beq
\tilde{L} - L\approx \sum_{k=i_1,\dots,i_M}\frac{\partial L}{\partial m_{k}} - \sum_{k'=j_1,\dots,j_M}\frac{\partial L}{\partial m_{k'}}.
\eeq

Denote the index among $i_{1},\dots,i_M$ with the maximum loss gradient by $i_{\max}$, and the index among $j_{1},\dots,j_M$ with the minimum loss gradient by $j_{\min}$.
Since $i_{\max}$ is selected while $j_{\min}$ is aborted during the score update, it can be derived that $\tilde{s}_{i_{\max}}>\tilde{s}_{j_{\min}}$ and ${s}_{i_{\max}}<{s}_{j_{\min}}$.
Therefore, based on~\eqref{equ_s_update},
\begin{align}
    \tilde{s}_{i_{\max}} - {s}_{i_{\max}}&>\tilde{s}_{j_{\min}}-{s}_{j_{\min}}
\Leftrightarrow   \frac{\partial L}{\partial m_{i_{\max}}} < \frac{\partial L}{\partial m_{j_{\min}}},
\end{align}
\beq
\Rightarrow \tilde{L} - L < M\cdot (\frac{\partial L}{\partial m_{i_{\max}}} - \frac{\partial L}{\partial m_{j_{\min}}}) < 0. \tag*{$\hfill \blacksquare$}
\eeq
 
\end{appendices}
 
\bibliographystyle{IEEEtran}
\bibliography{ref}

\end{document}